%% file: neurips_2025.tex
\documentclass{article}

\PassOptionsToPackage{numbers, sort&compress}{natbib}

\usepackage[final]{neurips_2025}

\usepackage[utf8]{inputenc} %
\usepackage[T1]{fontenc}    %
\usepackage{xcolor}         %
\definecolor{citecolor}{HTML}{0071bc}
\usepackage[pagebackref=true,breaklinks=true,colorlinks,bookmarks=false,citecolor=citecolor]{hyperref}
\usepackage{url}            %
\usepackage{booktabs}       %
\usepackage{amsfonts}       %
\usepackage{nicefrac}       %
\usepackage{microtype}      %

\input{preamble}

\newcommand{\papertitle}{Learning Skill-Attributes for\\Transferable Assessment in Video}
\title{\papertitle}

\author{%
  Kumar Ashutosh \\
  Univeristy of Texas at Austin\\
  \And
  Kristen Grauman \\
  University of Texas at Austin \\
}

\begin{document}

\maketitle

\vspace*{-0.22in}
\input{sec/0_abstract}

\vspace*{-0.2in}
\input{sec/1_introduction}

\input{sec/2_related}

\input{sec/3_method}

\input{sec/4_expts}

\input{sec/5_conclusion}

{
    \small
    \bibliographystyle{ieeenat_fullname}
    \bibliography{main}
}

\input{sec/X_supp}

\input{sec/Y_checklist}

\end{document}

%% file: preamble.tex
\usepackage{array}
\usepackage{colortbl}
\newcommand{\PreserveBackslash}[1]{\let\temp=\\#1\let\\=\temp}
\newcolumntype{C}[1]{>{\PreserveBackslash\centering}p{#1}}
\newcolumntype{L}[1]{>{\PreserveBackslash\raggedright}p{#1}}

\usepackage{graphicx}
\usepackage{pgfplots}
\pgfplotsset{compat=1.18}
\usepackage{subcaption}
\usepackage{tikz}
\newcommand{\modelname}{\textsc{CrossTrainer}}%

\newcommand{\KACr}[1]{{\color{black}#1}} 
\newcommand{\KGCR}[1]{{\color{black}{#1}}}

\definecolor{Gray}{gray}{0.95}
\definecolor{DarkGray}{gray}{0.55}

\usepackage{tcolorbox}
\tcbuselibrary{breakable} %

%% file: sec/0_abstract.tex
\begin{abstract}

Skill assessment from video entails rating the quality of a person's physical performance and explaining what could be done better.  Today's models specialize for an individual sport, and suffer from the high cost and scarcity of expert-level supervision across the long tail of sports.  Towards closing that gap, we explore transferable video representations for skill assessment.  Our \modelname~approach discovers \emph{skill-attributes}---such as balance, control, and hand positioning---whose meaning transcends the boundaries of any given sport, then trains a multimodal language model to generate actionable feedback for a novel video, e.g., \emph{``lift hands more to generate more power''} as well as its proficiency level, e.g., \emph{early expert}.  We validate the new model on multiple datasets for both cross-sport (transfer) and intra-sport 
 (in-domain) settings, where it achieves gains up to 60\% relative to the state of the art.  By abstracting out the shared behaviors indicative of human skill, the proposed video representation generalizes substantially better than an array of existing techniques, enriching today's multimodal large language models. Project page: \href{https://vision.cs.utexas.edu/projects/CrossTrainer/}{https://vision.cs.utexas.edu/projects/CrossTrainer/}.

\end{abstract}

%% file: sec/1_introduction.tex
\section{Introduction}
\label{sec:intro}
\vspace*{-0.05in}

The basis of assessing skilled physical activities, particularly sports, is largely visual.  Precisely how a tennis player grasps and swings their racquet; how a basketball player releases the ball to shoot a free throw; how a rock climber stretches and pulls to traverse the boulder---such visual details are discernible to the expert eye and essential for providing meaningful coaching.  Advances in multimodal video understanding could, therefore, transform AI-assisted coaching and skill assessment.  For example, future AI agents could provide personalized feedback to users based on videos captured on their phone or smartglasses, greatly expanding the accessibility of 1-1 coaching.  Similarly, AI could analyze how multiple players' skills would complement each other when building a team, or even detect patterns in injuries as a function of execution style.

\begin{figure}[t]
    \centering
    \includegraphics[width=\linewidth]{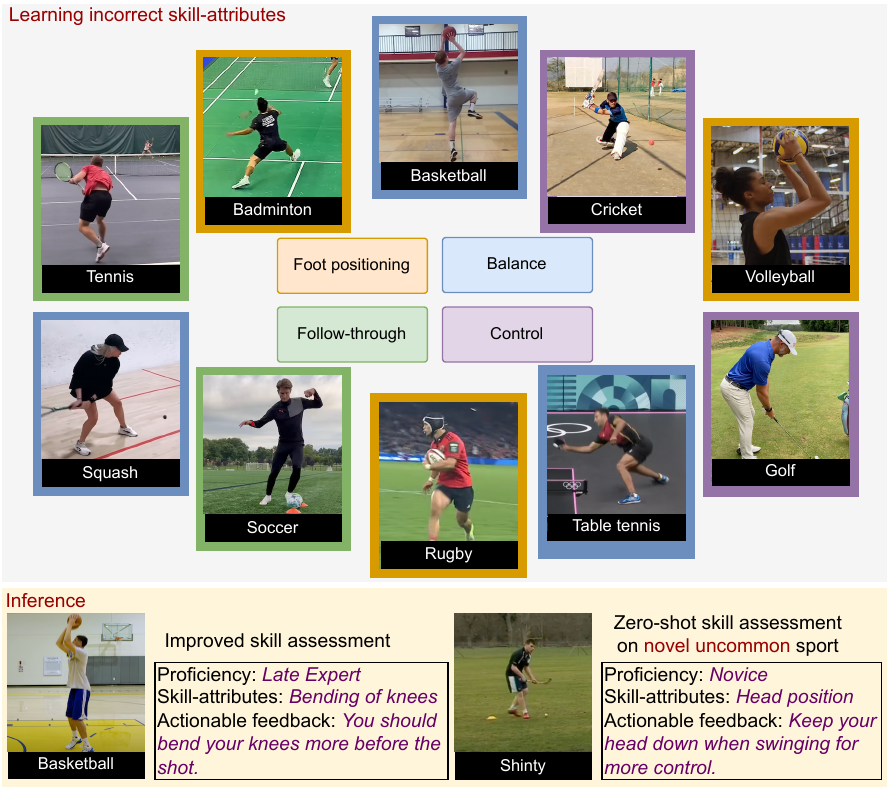}
    \caption{\textbf{Overview of the idea.} Given a short video of an athletic skill, what could be better?  Given video demonstrations from multiple sports, we learn skill-attributes that are incorrectly demonstrated, e.g., wrong foot positioning in badminton (\KGCR{top}). These skill-attributes are common across various sports, and transfer to novel uncommon sports, e.g., shinty (\KGCR{bottom}). %
    Our method improves both the in-domain and zero-shot settings. Sports chosen for illustration; see Sec.~\ref{sec:expts} for dataset details. 
    } \vspace{-0.5cm}
    \label{fig:teaser}
\end{figure}

Towards achieving the above, the core vision and machine learning task is as follows: given a video of an athlete's performance, estimate their skill level~\cite{action-quality-assessment,baller-gedas,stl-vs-mtl,interpretable-feedback,basket-gedas,egoexo4d,ego-exo-learn,whos-better-dima} and indicate what could be improved~\cite{qevd-panchal,expertaf,video-action-difference}. 

This task raises important unsolved challenges.  First of all, assessing %
skill requires \emph{fine-grained} understanding of all physical aspects.  Whereas traditional action understanding emphasizes high-level semantics~\cite{kinetics,kinetics-600,vit,mil-nce,videoclip,univl,internvideo2,internvideo}---and thus seeks \emph{invariance} to execution differences---here the subtle differences are exactly what matters.   For example, a novice and expert shooting a soccer ball into the net might accomplish the same overall action, but details about their approach, footwork, and trajectory of the ball are essential to analyze skill.
Secondly, compounding the challenge, 
supervision is costly and difficult to scale across the \emph{long tail} of sports.  An estimated 8,000 unique sports exist today across the world, but only a small subset is widely recorded and distributed---often driven by geographic and economic factors.  Similarly, expert supervision %
is expensive %
and cannot be scaled up easily using traditional annotator pools.  Making it worse, due to the common assumption that the axes of evaluation vary so wildly between sports, all prior work trains and tests on \emph{in-domain} data, i.e., the same drill, exercise, or sport~\cite{workout-form-assessment,baller-gedas,expertaf,action-quality-assessment,
group-aware-regression,quality-distribution-learning,egoexo4d,action-quality-assessment,stl-vs-mtl}.

We propose \modelname, a new approach to video-based skill analysis that accounts for these challenges.  Our key insight is to learn a universal \emph{skill-attribute} representation that transcends the boundaries of sports and hence allows sharing and transfer between them. 
A skill-attribute is a describable fine-grained concept about the physical performance that can take on different visual forms in different sports, e.g., \emph{ball control}, \emph{foot positioning}, \emph{coordination}, \emph{timing}, or \emph{balance}.  By automatically learning these shared properties from video-language commentaries, and then surfacing them during training, we aim to amplify the value of the limited training data available for any one sport.  

Building on the skill-attributes, we then train a generative multimodal language model to predict i) which skill-attributes are incorrectly demonstrated in a given video, ii) what the proficiency level is, and iii) free-form commentary about which adjustments would improve the performance.   Essentially we factor skill assessment into its generalizable cross-sport component (e.g., \emph{lacks control}) and its focused sports-specific component (e.g., \emph{``you should increase the spacing between your legs for better control of the soccer ball''}).  
Our formulation aims to unlock the translation of skill assessment from widespread, highly marketed sports to uncommon, low-resource sports, such as \emph{kabaddi}, \emph{frisbee}, \emph{waterpolo}, \emph{shinty}, \emph{kho-kho}, or \emph{bandy}. 
See Fig.~\ref{fig:teaser}.

Though this represents a departure from today's AI models~\cite{workout-form-assessment,baller-gedas,expertaf,action-quality-assessment,
group-aware-regression,quality-distribution-learning,egoexo4d,action-quality-assessment,stl-vs-mtl}, cognitive science supports 
both transferability across sports having similar skill sets and shared terminology for assessment~\cite{transferability-survey,cogsci-2,cogsci-invasive-non-invasive,parkour-as-transfer}.  For example, basketball players make better decisions playing soccer compared to tennis players, due to the former two sports' shared underlying properties~\cite{cogsci-2}; similarly, contact-sport athletes exhibit better zero-shot understanding of a new contact sport compared to non-contact sport athletes~\cite{cogsci-invasive-non-invasive}; and transferable skills arise from similar affordances~\cite{movement-transfer-1,movement-transfer-2} or environmental properties~\cite{parkour-as-transfer}.
Our work is the first to translate these intuitive findings from cognitive science into a working video understanding model.

We validate our ideas on three diverse datasets: Ego-Exo4D~\cite{egoexo4d}, which contains soccer, basketball, and rock climbing; QEVD~\cite{qevd-panchal}, which contains fitness exercises; and in-the-wild YouTube videos of people tutoring physical activities. 
We show that training an intermediate skill representation yields both superior \emph{in-domain} performance (for familiar sports) as well as better \emph{zero-shot} performance (where the particular sport or drill is never seen by the model during training).  
\modelname~outperforms all competitive baselines on all axes of assessment---actionable feedback, skill-attributes, and proficiency estimation---with relative gains up to $60\%$. 
Furthermore, compared to any of the baselines, \modelname~shows much more graceful degradation when transferring to a novel sport.
Overall our approach is a stepping stone for learning a unified skill-centric video representation for assessment and coaching in the presence of real-world data and annotation constraints.

%% file: sec/2_related.tex
\section{Related Work}
\label{sec:related}

\textbf{Learning representations from videos.} 
Substantial research explores new ways to learn video representations~\cite{videoclip,hero,hiervl,egovlp,egovlpv2},  targeting 
video understanding tasks of action recognition~\cite{omnivore,mvitv2,uniformer,memvit}, action anticipation~\cite{avt,rulstm,intention,whenwillyoudowhat,gao2017red}, procedural understanding~\cite{procedure-learning-fei-fei-li,procedure2,procedure3,bi2021procedure,naing2020procedure}, and temporal step localization~\cite{vina,unloc,vslnet,2d-tan,htstep-neurips2023}. 
Instructional videos offer a valuable window into skilled human activity~\cite{howto100m,coin,crosstask}, and recent work explores how to navigate between related how-to's~\cite{detours,paprika,task_graph} and identify their differences~\cite{stepdiff}.
Whereas prior work explores activity-centric representations emphasizing the semantics of \emph{what} is being done, our problem requires capturing \emph{how} it is being done, which we show is essential for skill assessment.

\textbf{Video-based sports analytics.} Sports analytics and assessment
raises a number of interesting challenges for computer vision~\cite{egoexo4d,action-quality-assessment,baller-gedas,stl-vs-mtl,interpretable-feedback,quality-distribution-learning,group-aware-regression}. Multiple ongoing workshops and challenges~\cite{CVsports,OpenFAD,soccernet} offer tasks including ball spotting, foul recognition, and game state reconstruction.  Prior work on skill assessment either assigns a score (or equivalence class label) to
a video demonstration~\cite{egoexo4d,action-quality-assessment,baller-gedas,stl-vs-mtl,interpretable-feedback,basket-gedas}, optimizes a group contrastive score distribution~\cite{quality-distribution-learning,group-aware-regression}, or chooses the better of two demonstrations~\cite{baller-gedas,skill-in-videos}. 
Broadening the scope of skill assessment beyond scoring videos, ExpertAF~\cite{expertaf} aims to provide \emph{actionable feedback} in the form of natural language commentary, relevant video retrievals, and generated pose corrections. 
As discussed above, all the above existing work in video-based skill assessment is limited to in-domain testing, whereas we explore transfer \emph{across} sports, enabled by the proposed skill-attributes.  In addition, orthogonal to the transfer contribution, our results across three datasets representing 6 distinct sports and fitness activities and 30 distinct drills  raises the bar in breadth of validation compared to any prior skill assessment work.

\textbf{Zero-shot generalization and attributes.} Testing on novel classes and scenarios is crucial for real open-world settings. %
In one line of work, zero-shot generalization is enabled by shared multimodal representations learned from %
noisy vision-language data, benefiting 
image classification~\cite{clip,xian2018zero,metaclip}, action recognition~\cite{frozenintime,zs-action,videococa}, video to text and text to video retrieval~\cite{videoclip,internvideo,internvideo2,hiervl,s3d-g,howtocaption}, or 
image segmentation~\cite{zs-segmentation,zs-segmentation-2,zs-segmentation-3}.  In another line of work, \emph{attributes} are intermediate variables~\cite{attribute-akata-2020,sylvain2020,palatucci2009,larochelle2008,savarese2011,dinesh2014} that can express a new category even without training images (e.g., zebras are \emph{black and white} and \emph{striped}).
Though they share our motivation for shared representations, all of the existing methods focus on semantics (what) as opposed to dynamic execution (how), and none are directly applicable to skill assessment from video.
Ours is the first work to explore attributes for fine-grained activities in video and the first to demonstrate the relevance for skill feedback. %

%% file: sec/3_method.tex
\section{Method}
\label{sec:method}

We introduce the problem statement in Sec.~\ref{sec:problem-def}, followed by an overview of key datasets (Sec.~\ref{sec:datasets}), our idea to extract skill-attributes (Sec.~\ref{sec:dataset-creation}), the full model (Sec.~\ref{sec:model}), and training (Sec.~\ref{sec:training-and-inference}).

\subsection{Problem definition}
\label{sec:problem-def}

At inference time, given a short video clip $V \in \mathcal{D}_{te}$ from the test set $\mathcal{D}_{te}$, we want to assess its quality, even if the exact same skill/drill was never seen in training---and even (more extreme) if the sport in $V$ was not seen during training. The desired assessment output covers three aspects: report the skill-attributes performed incorrectly, generate actionable feedback that can help the learner improve, and finally, estimate a proficiency score.  Consistent with the transfer observed in cognitive science studies~\cite{movement-transfer-1,movement-transfer-2,parkour-as-transfer}, we focus on physical skills executed by an individual and hence assume the video contains one person of interest; multi-player team interactions are also interesting but less amenable to transfer and outside the scope of this work.

To handle this task of assessing both in-domain and novel data, we employ a two-stage training process. In the pretraining stage, we train the model to generate the incorrectly demonstrated skill-attributes, i.e., we learn
a function $\mathcal{F}_a$ to predict the skill-\textbf{a}ttributes $\hat{S}=\{s_1, s_2, ...\}$ that the person in the video should improve on:
\begin{equation}
\mathcal{F}_a(V~|~\mathcal{D}_{tr} ) = \hat{S},
\end{equation}
where $\mathcal{D}_{tr}$ denotes the training set, e.g., if a person in the video is dribbling a soccer ball, $S$ can be \emph{control} and \emph{leg positioning}, supposing the person is incorrectly executing those two.  See Fig.~\ref{fig:teaser}, left.

In the second stage, we finetune the model for the remaining two aspects of assessment.
Firstly, we generate feedback conditioned on the inferred skill-attibute set $\hat{S}$, %
\begin{equation}
\mathcal{F}_t(V, \hat{S}~|~\mathcal{D}_{tr}) = T,
\end{equation}
where $T$ is a \textbf{t}extual actionable feedback statement for improvement, e.g., \emph{``the player is too straight and should bend more for a better control''}. While the skill-attributes are words or short phrases indicating suboptimal dimensions in general terms, the feedback consists of sentence(s) elaborating on what to fix in the context of the observed sport. 
Finally, we generate the \textbf{p}roficiency level of the person in $V$, again conditioned on $\hat{S}$:
\begin{equation}
\mathcal{F}_p(V, \hat{S}~|~\mathcal{D}_{tr}) = P,
\end{equation}
where $P$ is the proficiency class label, e.g., \emph{novice, intermediate, early expert} or \emph{late expert}.

All aspects of assessment are evaluated in both the \emph{in-domain} and \emph{zero-shot} settings, to explore both the absolute performance of our proposed approach with respect to the literature (in-domain) as well as its ability to transfer to novel sports and scenarios (zero-shot).  In the experiments we tackle multiple datasets of various sports and fitness activities, introduced next and detailed more in Sec.~\ref{sec:training-and-inference}.

\begin{figure}[t]
  \centering
  \begin{minipage}[b]{0.45\linewidth}
    \centering
    \includegraphics[width=\linewidth]{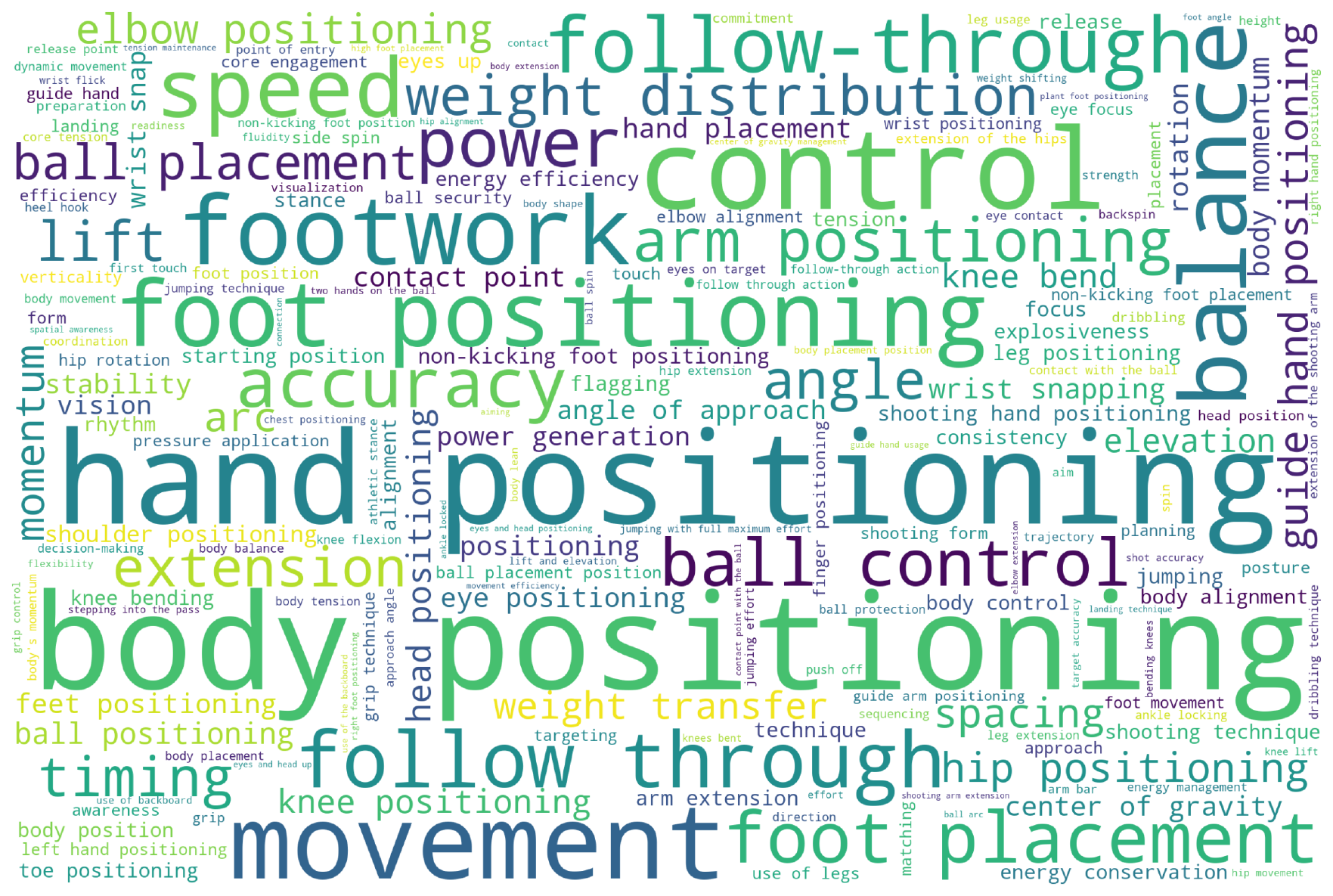}
  \end{minipage}
  \begin{minipage}[b]{0.45\linewidth}
    \centering
    \includegraphics[width=\linewidth]{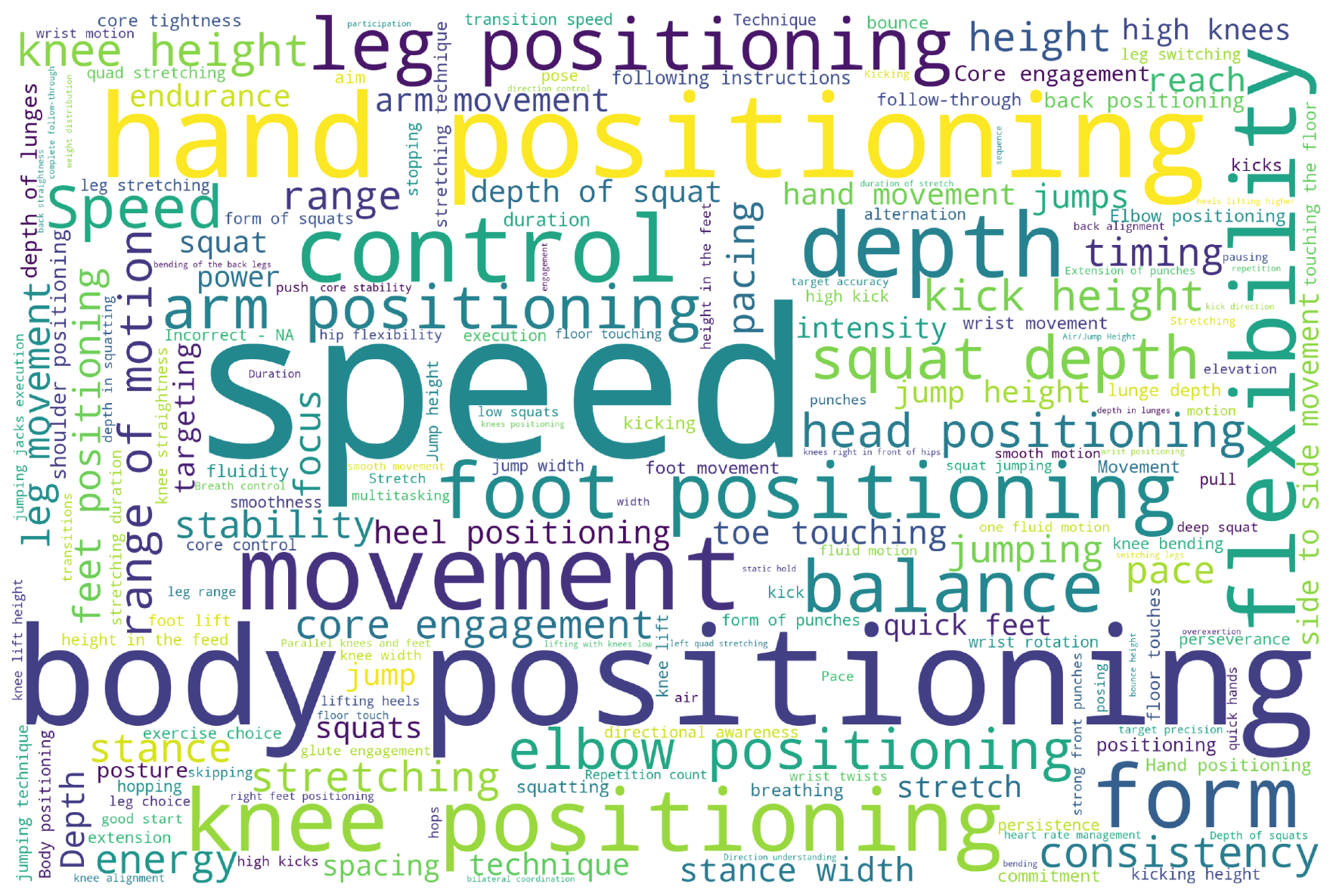}
  \end{minipage}
  \caption{\textbf{Discovered skill-attributes} from Ego-Exo4D~\cite{egoexo4d} (left) and QEVD~\cite{qevd-panchal} (right).  
  We see phrases reflecting generalizable physical concepts like control, hand/body positioning, and movement.
  }
  \label{fig:word-cloud}
\end{figure}

\subsection{Skilled physical activity datasets}\label{sec:datasets}

Before introducing our model, we next overview the key existing datasets leveraged in this study, since being familiar with their contents will help visualize our overall learning paradigm.
 \textbf{Ego-Exo4D}~\cite{egoexo4d} contains 2,593 videos, totaling 239 hours with 289 total participants playing 3 sports---soccer, rock climbing, and basketball. 
\textbf{Qualcomm Exercise Videos Dataset (QEVD)}~\cite{qevd-panchal} has 223 long home fitness exercise videos, totaling 13 hours, where 28 total participants perform 23 structured workout moves (like jumping jacks, planks, squats, etc.).  
To stretch our model on in-the-wild zero-shot settings, we also show qualitative results using \textbf{YouTube sports} videos collected by us---\emph{frisbee}, \emph{water polo} and \emph{soccer (juggling)}.

Both Ego-Exo4D and QEVD provide actionable feedback commentary for the videos, 
which consists of timestamped positive and negative critiques, ideas to correct form, pacing, or other aspects of execution.  Each sport is critiqued by expert coaches or players from that same sport. Given a video $V$, the expert pauses the video at multiple timepoints $t$ and provides verbal feedback, e.g., \emph{``...here he's showing good control but lacks speed, which is critical to an effective dribble...''}. Ego-Exo4D additionally has proficiency labels corresponding to four distinct skill levels, from novice to late expert. See Figure~\ref{fig:qual} and Supp.~for examples.

\subsection{Stage I: Discovering skill-attributes for pretraining supervision}
\label{sec:dataset-creation}

\begin{figure}[t]
  \centering
  \begin{minipage}[b]{0.95\linewidth}
    \centering
    \includegraphics[width=\textwidth]{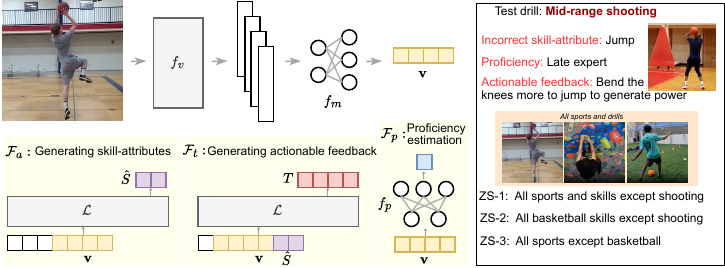}
  \end{minipage}
  \caption{\textbf{Method overview and evaluation settings.} (Left) We encode videos into tokens $\mathbf{v}$ that can be fed to a multimodal LLM $\mathcal{L}$, with a mapper $f_m$ that trains for skill-attributes. We use these visual tokens to generate skill-attributes (bottom left). Next, this pretraining is used to generate actionable feedback (bottom middle) and proficiency score (bottom right). (Right) Example of the various training settings for in-domain and zero-shot. 
  }
  \label{fig:model}
\end{figure}

To supervise the training of $\mathcal{F}_a$, we obtain the skill-attributes set $S$ for every video demonstration by sourcing it from Ego-Exo4D and QEVD commentaries---totaling around $34k$ unique feedback strings. 
 We use these commentaries as a signal to extract the skill-attributes the learner in the video should improve on, as detailed next. 
While the Ego-Exo4D and QEVD expert commentary represent existing video-language datasets nicely suited for our purposes, such commentary could potentially be sourced ``in the wild" from current video sharing platforms (e.g., Reddit, TikTok, etc.) where people provide similar verbal assessments for videos shared on social media. See Supp.

For each training sample, we prompt a large language model (LLM)~\cite{gpt4o} to extract skill-attributes that are suboptimally demonstrated according to its expert commentary text $T$.  We provide examples to help define the intent of skill-attributes to the LLM. See Supp.~for full prompt. This process yields $S=\{s_1, s_2, ...\}$ from the expert commentary $T$ at time $t$.
Note that $|S|$ can be of any length. %
Fig.~\ref{fig:word-cloud} shows word clouds of the discovered skill-attributes for the two video datasets. We observe that many salient phrases transcend sport boundaries, like \emph{body positioning}, \emph{balance}, \emph{control}, and \emph{movement}. 

Finally, we sample a video chunk around the time $t$, i.e., $[t-\mu_1, t+\mu_2]$ and associate it with the skill-attribute set $S$.\footnote{The commentary text has only a timepoint $t$, not a temporal extent; this window extraction is consistent with how the temporal ambiguity is handled in prior work~\cite{egovlp,naq,expertaf}.} %
By using an LLM combined with the raw commentary data, we obtain diverse, open-vocabulary skill-attributes for every training sample.  We draw on these resulting LLM-inferred annotations
when training the multimodal LLM (Sec.~\ref{sec:model}).  %

\subsection{Stage II: Skill assessment from video}
\label{sec:model}

Next we present our model design for learning $\mathcal{F}_a$.  Taking inspiration from the success of multimodal LLMs~\cite{llava,llava-med,videochat,video-chatgpt}, we convert a query video $V$ into visual tokens. Next, we input the visual tokens to a multimodal LLM, along with a prompt to predict the skill-attributes $S$. The output is parsed to create the prediction $\hat{S}$. Fig.~\ref{fig:model} shows the overview, and we describe each step next.

\textbf{Encoding video demonstrations.} We first use a standard video encoder $f_v$ 
(we use EgoVLP2~\cite{egovlpv2} and CLIP~\cite{clip}, though others are possible)
to extract video features from the video demonstration, $\mathbf{v}' = f_v(V)$, see Fig.~\ref{fig:model} (top left). Our model supports either single-view or multi-view inputs.  When available, multi-view observations of a skilled physical activity can give valuable detail; for example, an egocentric view of the basketball being shot by the hands along with one or more exocentric viewpoints of the player's full body pose would provide complementary information.  
When using multiple views, $\mathbf{v}'$ is simply the concatenation of features from individual views. The video spans $\mu_1+\mu_2$ seconds, as discussed in Sec.~\ref{sec:dataset-creation}, and one feature is extracted per second. Therefore, $\mathbf{v}'$ is a vector of dimension $(\mu_1+\mu_2) \times N$ where $N$ is the output feature dimension. %

Note that our representation is video-frame based, meaning the person's body pose is visible in the RGB but not explicitly extracted. While one could alternatively supplement the input with explicit body poses, our early experiments showed that the accuracy advantage is minimal compared to the high compute needed to get state-of-the-art poses~\cite{wham}, i.e., $34$ seconds for a $10$ second clip.

\textbf{Pretraining: Multimodal LLM for skill-attribute generation.} After encoding the videos, we input it to a multimodal LLM (M-LLM), denoted as $\mathcal{L}$. An LLM is a suitable choice since %
the output attributes are expressed in text format. We prompt the model as follows: \emph{``<video> Here is a video of a person doing <sport name>. Highlight up to <k> key concept areas where the person can improve: ...''}. We replace the \emph{`<video>'} tag with video encodings obtained above.

The output of the M-LLM contains `$k$' skill-attributes, distinct phrases that represent the dimensions of suboptimal execution in the provided video demonstration. These outputs are parsed to obtain the skill-attributes set $\hat{S}$, see Fig.~\ref{fig:model} (bottom left). 
To this end, we employ a trainable mapper $f_m$ that converts the feature vector $\mathbf{v'}$ to a representation compatible with the LLM, $\mathbf{v} = f_m(\mathbf{v}')$. The idea is that $f_v$ is a large pretrained model and kept frozen, while $f_m$ is trainable to convert the visual features to a multimodal representation.  
The supervising signal to train this M-LLM is the skill-attribute set $S$ obtained in Sec.~\ref{sec:dataset-creation}. We use standard log-likelihood loss~\cite{llama2,llama-herd,gpt4,llava}. \KACr{We emphasize that this pretraining enables the model to \emph{generate} skill-attributes, as opposed to retrieving from a closed set. We show the superiority of this approach over retrieval in the experiments.}

This pretraining stage results in aligning the video features $\mathbf{v}$ towards skill-attributes, thus making it suitable for the other axes of assessment---actionable feedback and proficiency estimation. We now use the pretraining weights for completing the assessment task suite, as described next.

\textbf{Skill-attributes for actionable feedback.}
For generating expert actionable feedback, we provide the output skill-attributes $\hat{S}$, along with video $V$, as input to the model $\mathcal{L}$ and output the actionable feedback. See Fig.~\ref{fig:model} (bottom middle). 
Next, we prompt the model $\mathcal{L}$ with a prompt as follows: \emph{``<video> Here is a video of a person doing <\KACr{sport name}>. Here are some possible axes that need improvement, as rated by an AI coach (may contain mistakes): <$\hat{S}$>. Give feedback on the execution that will help the person improve.''} As above, <video> is replaced by $\mathbf{v}$. Importantly, this prompt provides both $V$ and $\hat{S}$ as an additional guiding signal for the actionable feedback generation process. The generated actionable feedback $T$ contains ideas for improvement that are personalized to the video specifics, e.g., \emph{``you need to bend more while dribbling to maintain control''} as opposed to simply naming the skill dimension to improve, e.g., \emph{control}. %
While skill-attributes can be the same for two sports, the resulting actionable feedback will be geared towards the specific sport. This factoring helps make our model transferable, while also enabling sport-specific actionable feedback.

\textbf{Skill-attributes for proficiency estimation.}
Lastly, we %
deploy skill-attributes for proficiency estimation. %
As we want to discern how the skill-attributes representation compares to standard video features without such training, we employ a linear probe setting where we have a linear layer $f_p$ that takes in frozen $\mathbf{v}$ and outputs a class representing the proficiency $P$,
see Fig.~\ref{fig:model} (bottom right). The proficiency is a label of expertise: \emph{novice}, \emph{intermediate}, \emph{expert}, or \emph{late expert}. Only $f_p$ is trainable.

\subsection{Training and inference settings and implementation details}
\label{sec:training-and-inference}

\textbf{Train/test splits.} Our approach aims to improve both traditional in-domain and zero-shot skill assessment. 
The datasets organize video clips by their superclass sport and their subclass skill.  A skill is a drill or specific exercise, and each sport can have multiple skills. 
Ego-Exo4D has 3 superclasses and 5 subclasses: soccer has skills dribbling and penalty kick; basketball has skills Mikan layup, reverse layup, jump shot; rock climbing is not sorted into skills.  QEVD has 1 superclass (fitness) and 23 subclass skills (jumping jacks, squats, etc.).
To explore models' generalization ability,
we perform controlled experiments %
with the following train/test settings, in decreasing volume of available training data (see Fig.~\ref{fig:model} (right)):
\begin{itemize}
    \item \emph{Fully supervised (FS):} Train on all sports and skills, and test on held-out set of videos.  This represents in-domain testing.
    \item \emph{All sport zero-shot (ZS-1):} Train on all sports and skills, \emph{except} the target skill.  This means other skills from the same sport are seen during training.
    \item \emph{Familiar sport  zero-shot (ZS-2):} Train on all skills from the same sport, \emph{except} the target skill.  This means only the same sport is seen during training.
    \item \emph{Novel sport zero-shot (ZS-3):} Train only on $n-1$ sports, test on skills from the $n$-th unseen sport.  This means other skills from the same sport are \emph{not} seen during training.
\end{itemize}

\KACr{For each controlled experiment, we retrain the model to avoid any information leak between the train and the test splits, and between the seen and  novel sports.}

\begin{table}[t]\footnotesize
\centering
\caption{\textbf{Quantitative results.} Skill-attribute generation results (IoU@0.7) for Ego-Exo4D~\cite{egoexo4d} and QEVD~\cite{qevd-panchal} (top left). Actionable feedback generation results for Ego-Exo4D~\cite{egoexo4d} (top right) and QEVD~\cite{qevd-panchal} (bottom left). Metrics used to match SOTA on the corresponding dataset, B@4=BLEU@4, M=Meteor, R-L=ROUGE-L, B=BERT score. Proficiency estimation for individual sports (bottom right).  Standard errors are reported in text. 
}
\begin{minipage}{0.48\textwidth}
  \centering
\begin{tabular}{L{2.8cm}C{1.4cm}C{1.2cm}}
\toprule
\multicolumn{3}{c}{Skill-attribute generation} \\
\midrule
Method \hfill IoU@0.7 & Ego-Exo4D~\cite{egoexo4d} & QEVD~\cite{qevd-panchal} \\
\midrule
InternVideo2-NN \cite{internvideo2} & 14.0 & 23.8 \\
InternVideo2-FT \cite{internvideo2} & 15.0 & 24.5 \\
VideoChat2 \cite{videochat2} & 9.3 & 16.9 \\
LLaVA \cite{llava} & 9.7 & 17.3 \\
LLaVA-FT \cite{llava} & 14.6 & 26.9 \\
Stream-VLM~\cite{qevd-panchal} & 14.5 & 28.0 \\
ExpertAF~\cite{expertaf} & 15.0 & 28.1 \\
Attribute-Retrieval & 19.7 & 32.4 \\
\rowcolor{Gray}
\textbf{\modelname} & \textbf{25.7} & \textbf{37.6} \\
\bottomrule
\end{tabular}
\end{minipage}%
\hfill
\begin{minipage}{0.48\textwidth}
  \centering
\begin{tabular}{L{3.1cm}C{0.7cm}C{0.7cm}C{0.7cm}}
\toprule
\multicolumn{4}{c}{Actionable feedback on Ego-Exo4D~\cite{egoexo4d}} \\
\midrule
Method & B@4 & M & R-L \\
\midrule
InternVideo2-NN \cite{internvideo2} & 42.1 & 46.9 & 49.3 \\
InternVideo2-FT \cite{internvideo2} & 42.9 & 47.6 & 50.0 \\
VideoChat2 \cite{videochat2} & 27.8 & 44.3 & 41.9 \\
LLaVA \cite{llava} & 28.5 & 44.1 & 44.2 \\
LLaVA-FT \cite{llava} & 43.5 & 48.5 & 51.5 \\
LLaVA-FT w/ pose \cite{llava} & 43.6 & 48.5 & 51.7 \\
PoseScript/Fix \cite{posescript,posefix} & 24.1 & 44.5 & 46.3 \\
ExpertAF~\cite{expertaf,gpt4} & 44.9 & 49.6 & 54.6 \\
\rowcolor{Gray}
\textbf{\modelname} & \textbf{45.6} & \textbf{51.7} & \textbf{57.8} \\
\quad \quad w/o two-stage & 43.8 & 48.8 & 52.3 \\
\bottomrule
\end{tabular}
\end{minipage}

\begin{minipage}{0.48\textwidth}
  \centering
\begin{tabular}{L{3.1cm}C{0.7cm}C{0.7cm}C{0.7cm}}
\toprule
\multicolumn{4}{c}{Actionable feedback on QEVD~\cite{qevd-panchal}} \\
\midrule
Method & M & R-L & B \\
\midrule
Socratic-LLaMA-2-7B & 9.4 & 7.1 & 86.0 \\
Video-ChatGPT \cite{video-chatgpt} & 10.8 & 9.3 & 86.3 \\
LLaMA-VID \cite{llama-vid} & 10.6 & 9.0 & 86.0 \\
Stream-VLM \cite{qevd-panchal} & 12.7 & 11.2 & 86.3 \\

\rowcolor{Gray}
\textbf{\modelname} & \textbf{17.6} & \textbf{18.1} & \textbf{87.8} \\
\quad \quad w/o two-stage & 12.1 & 10.8 & 86.0 \\
\bottomrule
\end{tabular}
\end{minipage}
\hfill
\begin{minipage}{0.48\textwidth}
  \centering
\begin{tabular}{L{2.0cm}C{1.0cm}C{1.0cm}C{1.2cm}}
\toprule
\multicolumn{4}{c}{Proficiency estimation on Ego-Exo4D~\cite{egoexo4d}} \\
\midrule
Method & B.ball & Soccer & Rock Cl. \\
\midrule
EgoVLPv2 \cite{egovlpv2} & 48.0 & 62.5 & 34.0 \\
\rowcolor{Gray}
\textbf{\modelname} & \textbf{53.1} & \textbf{68.8} & \textbf{37.1} \\
\bottomrule
\end{tabular}
\end{minipage}%

\label{tab:merged-results}
\vspace{-0.4cm}
\end{table}

\textbf{Model architecture.} $f_v$ is EgoVLPv2~\cite{egovlpv2}  for Ego-Exo4D~\cite{egoexo4d} (precomputed with dataset) and CLIP~\cite{clip} for QEVD~\cite{qevd-panchal} (lightweight to extract). %
We use one feature per second with  a $4096$-d output.
For Ego-Exo4D~\cite{egoexo4d}, we use the ego and four exo views, while QEVD~\cite{qevd-panchal} is single-view. $f_m$ is a two layered MLP with GELU activation, consistent with \cite{improvedllava}. The MLP+GELU module takes in $4096$-d representation, consistent with the embedding dimension of the multimodal LLM $\mathcal{L}$, Llama-3.1-8B-Instruct~\cite{llama3modelcard}. $f_p$ is a linear layer with output size $4$, the number of proficiency levels.

\textbf{Training details.} We train both $\mathcal{F}_a$ and $\mathcal{F}_t$ in LoRA setting~\cite{lora}, with rank $128$, alpha $256$, and dropout $0.05$, for efficiency. The best performance is obtained with a learning rate of $2\times10^{-3}$ for $f_m$ and $2\times10^{-4}$ for $\mathcal{L}$. Recall that $f_v$ is kept frozen. The model is trained for $2$ epochs or till convergence. Total training time depends on the dataset setting, varying between $1-3$ hours. All experiments are performed on one GH200 NVIDIA node.

%% file: sec/4_expts.tex
\vspace*{-0.1in}
\section{Experiments and Results}
\label{sec:expts}
\vspace*{-0.05in}

\begin{figure}[t]
    \centering

    \includegraphics[width=\textwidth]{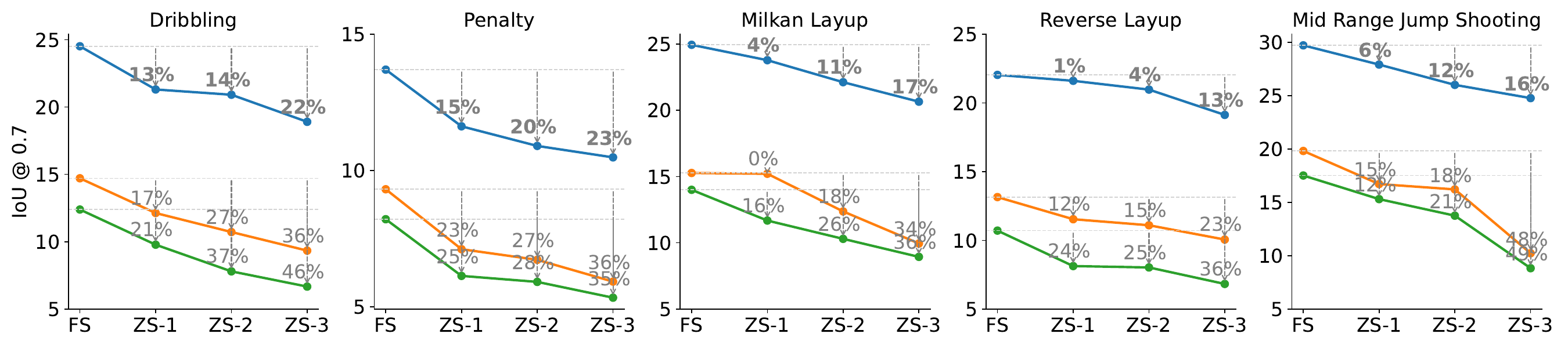}
    \includegraphics[width=\textwidth]{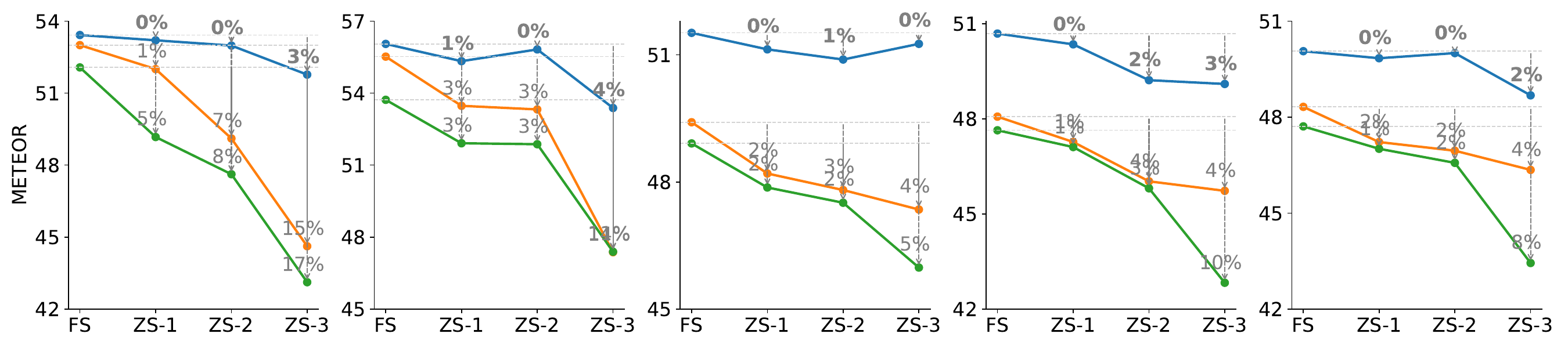}
    \caption{\textbf{Zero-shot performance.}  Performance trend when testing on various skills (dribbling, penalty, etc.) for different in-domain and zero-shot training settings (FS, ZS-1, etc.) for skill-attribute generation (top) and actionable feedback generation (bottom) for Ego-Exo4D. The relative drop in performance w.r.t. FS is shown as a percentage. Our method is consistently the best for all methods, and the relative drop is the least for all zero-shot variants (ideal curve would be flat and high). See Supp.~for QEVD. Legend: \textcolor[rgb]{0.1216,0.4667,0.7059}{\rule[0.5ex]{1.5em}{0.6pt}} \ \modelname, \textcolor{orange}{\rule[0.5ex]{1.5em}{0.6pt}} \ ExpertAF~\cite{expertaf} and \textcolor[rgb]{0.17,0.63,0.17}{\rule[0.5ex]{1.5em}{0.6pt}} \ InternVideo2~\cite{internvideo2}. 
    }
    \vspace{-0.2cm}
    \label{fig:zs-graphs}
\end{figure}

We first present actionable feedback generation (Sec.~\ref{sec:af-results}) for its focus in prior work and directly comparable SOTA methods. Next, we validate the skill-attributes pretraining (Sec.~\ref{sec:pretraining-results}). Finally, we present proficiency estimation (Sec.~\ref{sec:score-results}) and explore long-tail in-the-wild YouTube videos (Sec.~\ref{sec:youtube-results}).

\subsection{Generating actionable feedback}
\label{sec:af-results}

First we evaluate actionable feedback generation: given a video, generate the natural language commentary explaining what to correct. %

\textbf{Baselines.} We compare against the SOTA ExpertAF~\cite{expertaf} and Stream-VLM~\cite{qevd-panchal} models as well as the original baselines from their respective experiments, which include strong multimodal LLMs with model size 7B-8B.  \modelname~has fewer trainable parameters due to our use of LoRA~\cite{lora}, but we leave the baselines in their full (non-LoRA) form to report their most accurate numbers. \KACr{``w/o two-stage'' is the end-to-end ablation not using skill-attributes.}

\textbf{Metrics.} Following ~\cite{expertaf,qevd-panchal}, we report language metrics---METEOR~\cite{meteor}, BLEU-4~\cite{bleu}, ROUGE~\cite{rouge} and BERT-score~\cite{bert}, reported out of 100; 
higher is better. %
While establishing ground truth commentary is nuanced and there could be multiple valid feedback statements for a given video, prior work~\cite{expertaf} shows that this evaluation paradigm correlates strongly with human subjects' evaluation of the generated commentary.

\textbf{Fully supervised in-domain results.} 
 Tab.~\ref{tab:merged-results} (top right and bottom left) 
 shows the results.  \modelname~clearly outperforms all prior work and strong baselines. 
 Standard error is less than $0.1$ for all metrics. This result clearly supports using skill-attributes as an intermediate representation for  actionable feedback. \KACr{We investigate the robustness of actionable feedback generation to the choice of the LLM $\mathcal{L}$ and the noise level in the Supp}. Fig.~\ref{fig:qual} (first two rows) shows qualitative outputs for both datasets, where our model correctly captures the expert's feedback.

\textbf{Zero-shot transfer.} Fig.~\ref{fig:zs-graphs} (bottom row) 
shows the trend when transferring the knowledge from one domain to another in Ego-Exo4D~\cite{egoexo4d} (see Supp.~for QEVD~\cite{qevd-panchal}; results are similar). We plot the best M-LLM (ExpertAF~\cite{expertaf}) and retrieval (InternVideo2-FT~\cite{internvideo2}) baselines; we see a similar trend in all other baselines.  
Our method obtains the best performance for all training settings and degrades most gracefully as training data coverage diminishes in the increasingly difficult zero-shot settings (ZS-1, ZS-2, ZS-3).
Our max drop is $4\%$,  vs.~$17\%$ for the baseline. 
In Fig.~\ref{fig:qual} (bottom left), we show a confusion matrix denoting better transfer from soccer to basketball, and vice versa, compared to rock climbing---reinforcing observations in cognitive science~\cite{transferability-survey,movement-transfer-1,movement-transfer-2}.
In summary, \modelname~demonstrates robust to transfer to novel sports and drills.

\subsection{Learning skill-attributes in pretraining}
\label{sec:pretraining-results}

\begin{figure}[t]
  \centering
  \includegraphics[width=\textwidth]{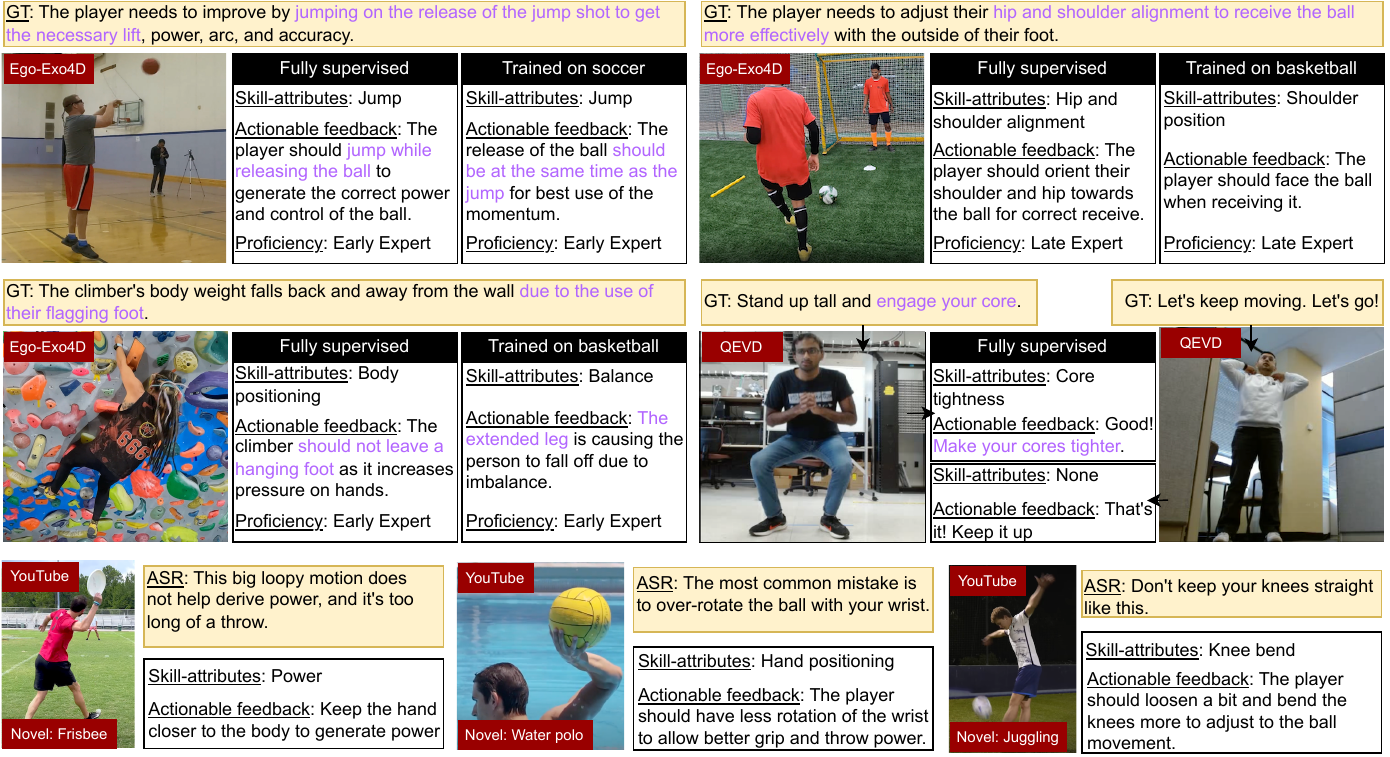}
  \includegraphics[width=0.46\textwidth]{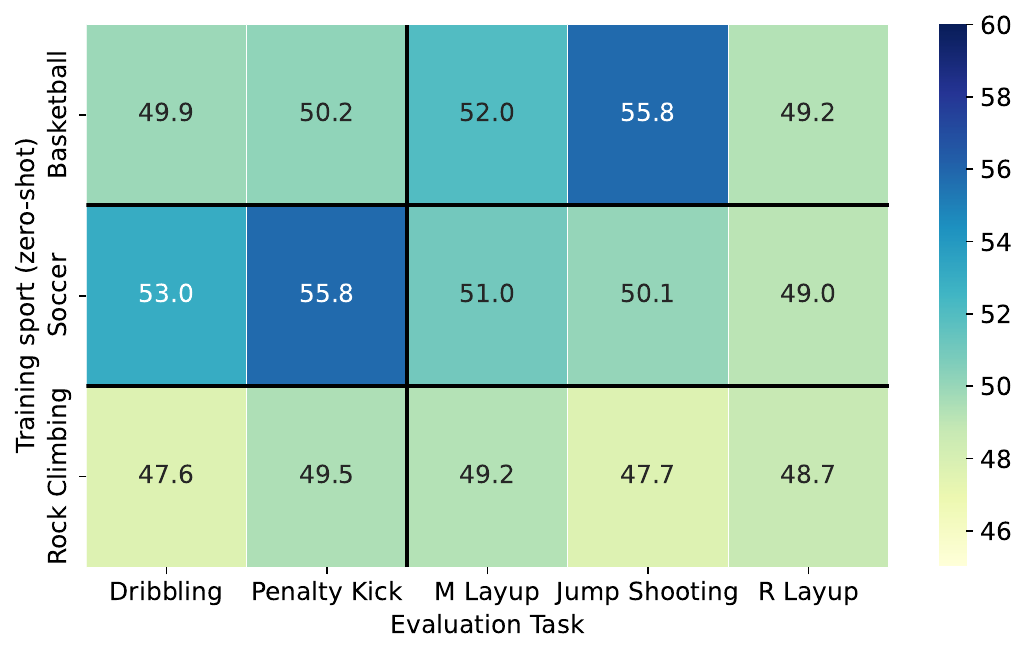}\hfill  \includegraphics[width=0.48\textwidth]{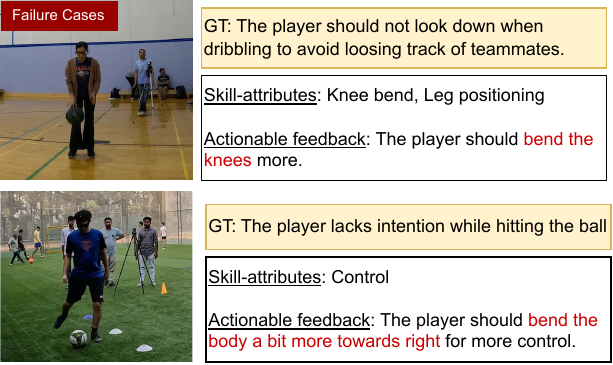}
  \caption{\textbf{Qualitative results.} \modelname~generates skill-attributes, actionable feedback, and proficiency for samples from both Ego-Exo4D~\cite{egoexo4d} and QEVD~\cite{qevd-panchal}. The outputs are meaningful even in the zero-shot setting (first two rows). Our method is also applied to in-the-wild videos from YouTube with novel sports (frisbee and water polo) and even new drills (juggling in soccer) with feedback matching the YouTube expert's comments (transcribed with ASR) (third row). Confusion matrix  shows better transfer between related sports (bottom left). \textbf{Failure cases} here and in Supp.~show the difficulty of the task, especially non-visual feedback like \emph{lacking intent} (bottom right). 
  }
  \label{fig:qual}
  \vspace{-0.5cm}
\end{figure}

Next we evaluate the skill-attribute prediction (c.f.~Sec.~\ref{sec:dataset-creation}). %

\textbf{Baselines.}
We 
compare to retrieval baselines (InternVideo2~\cite{internvideo2}, nearest neighbor (NN) and finetuned (FT), and Attribute-Retrieval trained with contrastive learning), zero-shot multimodal baselines (VideoChat2~\cite{videochat2}, LLaVA~\cite{llava}), and a finetuned version of the best multimodal baseline (LLaVA-FT). 
In addition, we again compare to ExpertAF~\cite{expertaf}, a SOTA model for generating actionable feedback, but here we convert its output to skill attributes by text-only LLM prompting~\cite{gpt4o}.  
All the baselines have access to the same actionable feedback data as ours~\cite{egoexo4d,qevd-panchal}, and they directly supervise with it.

\textbf{Metrics.} Since skill-attributes are a list of phrases, we employ IoU-based matching (a.k.a. Jaccard index)~\cite{jaccard-1,jaccard-2} between the inferred annotations (Sec.~\ref{sec:dataset-creation}) and
generated skill-attributes. We call a pair a match if BERT-score %
is more than $k$, and report IoU$@k$ for k=0.7 here and $\{0.8, 1.0\}$ in Supp.

\textbf{Fully supervised in-domain results.}  
Tab.~\ref{tab:merged-results} (top left) shows the results for Ego-Exo4D~\cite{egoexo4d} and QEVD~\cite{qevd-panchal}. 
\modelname~outperforms all zero-shot and trained baselines that \emph{post process} actionable feedback to obtain skill-attributes, by more than $10\%$. Moreover, our method outperforms Attribute-Retrieval by $6\%$---showcasing the effectiveness of our generative approach. 
The standard error is less than $0.5$ for all metrics and all methods. Fig.~\ref{fig:qual} shows examples predicting the skill-attributes that the person should improve.

\textbf{Zero-shot transfer.} Fig.~\ref{fig:zs-graphs} (top row of plots) shows the zero-shot transfer %
results for Ego-Exo4D~\cite{egoexo4d} (see Supp. for QEVD~\cite{qevd-panchal}; results are similar). 
As above, we see a clear advantage of our training scheme.
Moreover, \modelname~declines much more gracefully than the baselines. 
Overall, these results show the effectiveness of our skill-attribute guided pretraining for zero-shot transfer.

\subsection{Estimating demonstrator proficiency}
\label{sec:score-results}

Next we evaluate the quality of the learned video representation $\mathbf{v}$ for proficiency estimation.

\textbf{Baselines and metrics.} We compare the strength of the video representation with respect to the frozen representation $f_v$, i.e., EgoVLPv2~\cite{egovlpv2} for all scenarios. We report classification accuracy.

\textbf{Results.} Tab.~\ref{tab:merged-results} (bottom right) 
shows the results. We see a clear gain of up to $6\%$ when using the pretrained video representation, with standard error <1. Fig.~\ref{fig:qual} shows some example predictions, where our model can distinguish between \emph{early} and \emph{late expert}.\footnote{Note that QEVD does not provide proficiency labels.} This experiment suggests another potential use case in pretraining skill-centric representations for better assessment and feedback.

\subsection{Testing long tail in-the-wild YouTube sports videos}
\label{sec:youtube-results}

Finally, we explore applying \modelname~to
novel sports and drills in in-the-wild videos from YouTube. We extract $10$ clips from tutorial videos where an expert tutor demonstrates the incorrect way of doing a drill, while also explaining the incorrectness \KGCR{(i.e., the target feedback, which is withheld from our model)}. \KACr{Details of obtaining the videos from YouTube are given in the Supp}. We use the expert's transcribed ASR text to compare with the output of our model.   

\textbf{Results.} Fig.~\ref{fig:qual} (fourth row) shows example results (rest in Supp.). We see that a model trained on Ego-Exo4D~\cite{egoexo4d} videos (soccer, basketball, rock climbing), is able to predict the issue with demonstrations in novel long-tail sports frisbee and water polo. Moreover, we also see correct feedback in a novel juggling drill in soccer, which \modelname~is not trained with. We observe that learning a sport-specific vocabulary is difficult in zero-shot transfer. Nonetheless, the essence is captured and described in text, e.g., even though the model does not understand the \emph{loopy motion}, it understands that keeping the hands closer to the body helps in generating more power---a fact known and applied in various physical scenarios.  \KACr{We also verify the generations with a user study. Human subjects not associated with the project rated $75\%$ of the generations as actionable and correct. Every generation is \KGCR{judged} by three raters, and we take a majority vote.}

Failure cases in Fig.~\ref{fig:qual} (bottom right) showcase the difficulty in capturing subtle mistakes, especially non-visual feedback like \emph{intent}, \emph{decision}. We \KGCR{further} discuss limitations and societal impact in Supp.

%% file: sec/5_conclusion.tex
\vspace*{-0.1in}
\section{Conclusion}
\vspace*{-0.05in}

 We introduced \modelname---a novel approach that discovers \emph{skill-attributes} from video demonstrations. These skill-attributes are generalizable, enabling a zero-shot transfer of skill assessment to novel sports and drills. Our experiments show notable gains in actionable feedback generation and proficiency estimation for both in-domain and zero-shot settings.
In the future, we will explore ways to quantify sport relatedness to predict the transferability between sports, as well as explicit representations to capture the environmental context.

\section*{Acknowledgements}

Thanks to the anonymous NeurIPS reviewers for their valuable feedback.  
This research is supported in part by the UT Austin IFML AI Institute.  Compute is from the Vista GPU Cluster through the Center for Generative AI (CGAI) and the Texas Advanced Computing Center (TACC) at the University of Texas at Austin.

%% file: sec/X_supp.tex
\newpage

\appendix

\section{\emph{Supplementary material for} \papertitle}

\subsection{Table of content}

This supplementary contains the following:

\begin{itemize}
    \item \textbf{A supplementary video} that first motivates the problem with some example actionable feedback requests from learners on Reddit (discussed in Sec.~\ref{sec:dataset-creation}). Next, we show qualitative results from Ego-Exo4D~\cite{egoexo4d}, QEVD~\cite{qevd-panchal}, and YouTube videos. Finally, we also show some failure cases.

    \item \textbf{Sec.~\ref{supp:llm}: LLM prompt} that we use to obtain skill-commentary from expert commentary.

    \item \textbf{Sec.~\ref{supp:quant}: Additional quantitative} results to show  metrics for skill-attribute generation, and QEVD~\cite{qevd-panchal} zero-shot results.
    \item We discuss \textbf{limitations} in \textbf{Sec.~\ref{supp:limitations}}.
    \item We also discuss \textbf{societal impact} in \textbf{Sec.~\ref{supp:societal}}.
\end{itemize}

\subsection{LLM prompt for obtaining skill-attribute from expert commentary}
\label{supp:llm}

In Sec.~\ref{sec:dataset-creation}, we discuss using a large language model (LLM) to extract skill-attributes that are suboptimally performed. We use the following prompt:

\vspace{0.25cm}

\begin{tcolorbox}[breakable, boxrule=0.2mm]
\textbf{System:} Answer the question regarding a commentary about a sports drill. Do not add information not present in the question.

 \textbf{User:} The transcript of an expert commentating on a \textsc{Sport name comes here} drill is given below. List down the concepts that are correct and incorrect in the drill, as noted by the expert. The concepts are distinct aspects of the skill, e.g., control, body positioning, speed, body movement, hand position, and so on. Feel free to come up with newer concepts and write the response in two lines. The first line should contain the correctly shown concepts, and the second line should contain the incorrectly shown concepts. It should be in this format. 
 
 Correct - comma separated concepts.
 
 Incorrect - comma separated concepts.
 
 Here is the expert feedback:

 \textsc{Narration comes here}

 \textbf{Assistant:} 

\end{tcolorbox}
\vspace{0.25cm}

We prompt the LLM to provide both correctly and incorrectly demonstrated skill-attributes. Our Attribute-Retrieval baseline (Sec.~\ref{sec:score-results}) uses both of them.

\begin{table}[t]\footnotesize
\centering
\caption{Buckets of exercises in QEVD~\cite{qevd-panchal} grouped by similarity in execution and effect.}
\vspace{0.2cm}
\label{tab:exercise-groups}
\begin{tabular}{|p{3cm}|p{4cm}|p{5cm}|}
\hline
\textbf{Group name}& \textbf{Exercises} & \textbf{Remark} \\
\hline
Stretches \& mobility & quad stretch, armcrosschest, good morning beginner, floor touches, toe touchers & Focused on flexibility and range of motion; often used in warm-up or cooldown phases. Involves static or slow dynamic movement. \\
\hline
Cardio \& agility & high knees, quick feet, jumping jacks, air jump rope, butt kickers, puddle jumps & Elevates heart rate with low to moderate resistance; emphasizes agility and coordination with repetitive footwork. \\
\hline
Leg strength \& lower-body & squats, squat jumps, squat kicks, walking lunges, lunge jumps, standing kicks & Targets glutes, quads, hamstrings through controlled or explosive leg movements. Builds strength and endurance. \\
\hline
Core \& upper-body & plank taps, moving plank, pushups, shoulder gators & Focuses on core stabilization and upper body strength, particularly arms, shoulders, and chest. Often bodyweight-based. \\
\hline
Full-body & boxing squat punches, mountain climbers & High-intensity, compound movements that engage multiple muscle groups while promoting coordination and rhythm. \\
\hline
\end{tabular}
\vspace{0.2cm}

\end{table}

\begin{figure}[t]
    \centering

    \includegraphics[width=\textwidth]{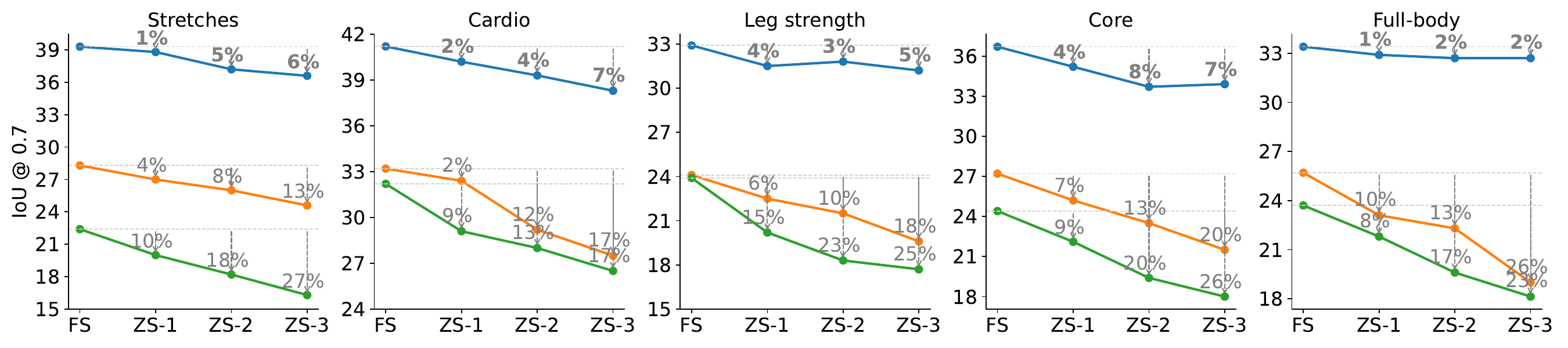}
    \includegraphics[width=\textwidth]{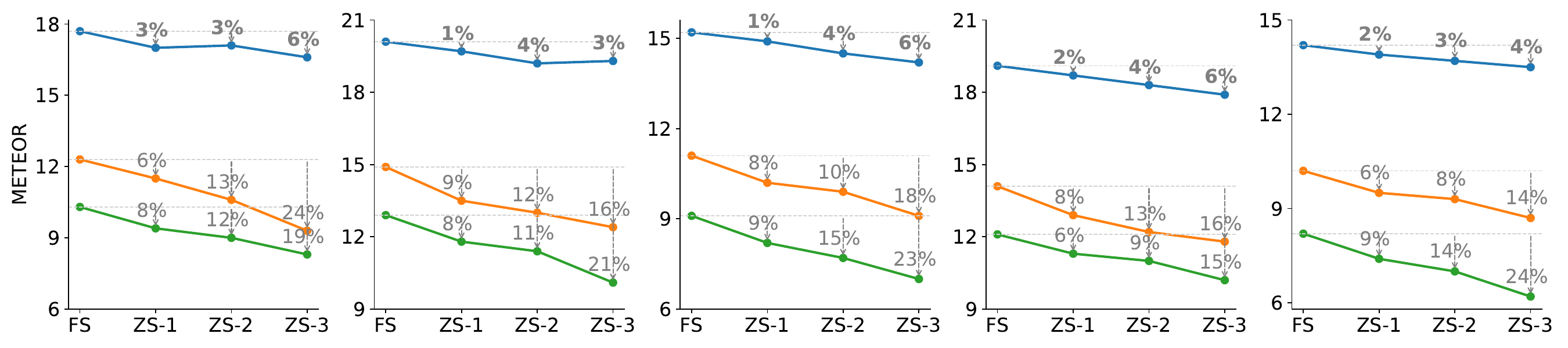}
    \caption{\textbf{Zero-shot performance.}  Performance trend when testing on various skills (stretches, cardio, etc.) for different in-domain and zero-shot training settings (FS, ZS-1, etc.) for skill-attribute generation (top) and actionable feedback generation (bottom) for QEVD~\cite{qevd-panchal}. Legend: \textcolor[rgb]{0.1216,0.4667,0.7059}{\rule[0.5ex]{1.5em}{0.6pt}} \ \modelname, \textcolor{orange}{\rule[0.5ex]{1.5em}{0.6pt}} \ Stream-VLM~\cite{qevd-panchal} and \textcolor[rgb]{0.17,0.63,0.17}{\rule[0.5ex]{1.5em}{0.6pt}} \ InternVideo2~\cite{internvideo2}. 
    }
    \vspace{-0.2cm}
    \label{fig:zs-qevd}
\end{figure}

\KACr{
\subsection{Obtaining YouTube videos for testing long tail in-the-wild sports and drills}

We evaluate the performance of our model on zero-shot sports and drills from in-the-wild YouTube videos in Sec.~\ref{sec:youtube-results}. To obtain a dataset for this test, we create a list of novel sports and novel drills within the Ego-Exo4D~\cite{egoexo4d} and QEVD~\cite{qevd-panchal} sports (basketball, soccer, rock climbing, exercise), and search for videos on YouTube with their coaching videos. Our criteria is that the video should contain a mistake done by a learner, and a coach giving feedback. Note that many videos show only the correct demonstration; hence, obtaining videos for our task is challenging.

Specifically, we randomly selected some common sports—soccer (juggling), basketball (dribbling) and some rarer ones like water polo, korfball, polo, jai alai, kin ball, frisbee. We search more than $50$ videos with keywords like ``Coaching video of <sport/drill>'', ``<sport/drill> training session for beginners'', ``<sport/drill> common mistakes for beginners'', ``Dos and don’ts in <sport/drill>''. The videos are manually watched to find the desired coaching instances. The process took overall 12 hours, and was done by graduate students not associated with this project. 

}

\begin{table}[t]\footnotesize
\centering
\caption{\textbf{Quantitative results.} Skill-attribute generation results for IoU$@k$ for $k\in \{0.7, 0.8, 1.0\}$ for Ego-Exo4D~\cite{egoexo4d} and QEVD~\cite{qevd-panchal}, extension of Tab.~\ref{tab:merged-results} (top left) on remaining $k$ values. 
}
\vspace{0.1cm}
\begin{minipage}{0.48\textwidth}
  \centering
\begin{tabular}{L{3.1cm}C{0.7cm}C{0.7cm}C{0.7cm}}
\toprule
Method \hfill IoU & @0.7 & @0.8 & @1.0 \\
\midrule
InternVideo2-NN \cite{internvideo2} & 14.0 & 8.9 & 7.7 \\
InternVideo2-FT \cite{internvideo2} & 15.0 & 9.4 & 8.2 \\
VideoChat2 \cite{videochat2} & 9.3 & 6.6 & 4.0 \\
LLaVA \cite{llava} & 9.7 & 7.2 & 4.9 \\
LLaVA-FT \cite{llava} & 14.6 & 9.1 & 8.1 \\
Stream-VLM~\cite{qevd-panchal} & 14.5 & 9.1 & 8.3 \\
ExpertAF~\cite{expertaf} & 15.0 & 9.5 & 8.4 \\
Attribute-Retrieval & 19.7 & 12.7 & 10.7 \\
\rowcolor{Gray}
\textbf{\modelname} & \textbf{25.7} & \textbf{15.9} & \textbf{14.4} \\
\bottomrule
\end{tabular}
\end{minipage}%
\hfill
\begin{minipage}{0.48\textwidth}
  \centering
\begin{tabular}{L{3.1cm}C{0.7cm}C{0.7cm}C{0.7cm}}
\toprule
Method \hfill IoU & @0.7 & @0.8 & @1.0 \\
\midrule
InternVideo2-NN \cite{internvideo2} & 23.8 & 16.3 & 14.8 \\
InternVideo2-FT \cite{internvideo2} & 24.5 & 16.6 & 15.3 \\
VideoChat2 \cite{videochat2} & 16.9 & 11.4 & 10.1 \\
LLaVA \cite{llava} & 17.3 & 12.5 & 11.8 \\
LLaVA-FT \cite{llava} & 26.9 & 19.2 & 18.2 \\
Stream-VLM~\cite{qevd-panchal} & 28.0 & 19.9 & 18.6 \\
ExpertAF~\cite{expertaf} & 28.1 & 19.7 & 18.3 \\
Attribute-Retrieval & 32.4 & 24.7 & 23.0 \\
\rowcolor{Gray}
\textbf{\modelname} & \textbf{37.6} & \textbf{29.8} & \textbf{28.1} \\
\bottomrule
\end{tabular}
\end{minipage}

\label{tab:sa-complete}
\end{table}

\KACr{

\begin{table}[t]\footnotesize
\centering
\caption{\textbf{Robustness and sensitivity of LLM:} Comparison of actionable feedback generation with various sources of skill-attributes (left). Comparison of BERT similarity score of skill-attributes generated by our method with skill-attributes obtained from different LLMs and prompts (right).}
\vspace{0.1cm}

\begin{minipage}{0.64\textwidth}
    \begin{tabular}{l l c c c}
\toprule
\textbf{Skill-attrib. source} & \textbf{Test-set $\mathcal{L}$} & \textbf{B@4} & \textbf{M} & \textbf{R-L} \\
\midrule
Llama-3 8B (orig) & Llama-3 8B (orig) & 45.6 & 51.7 & 57.8 \\
Mistral 8B~\cite{mistral8b} & Llama-3 8B (orig) & 45.3 & 51.8 & 57.5 \\
Llama-3 8B (orig) & Mistral 8B & 45.8 & 51.8 & 57.8 \\
\midrule
Llama-3 8B (orig) &&&& \\
\quad \quad w/ 10\% noise & Llama-3 8B (orig) & 45.2 & 50.5 & 56.2 \\
\quad \quad w/ 20\% noise & Llama-3 8B (orig) & 45.1 & 50.0 & 54.9 \\
\quad \quad w/ 30\% noise & Llama-3 8B (orig) & 44.3 & 49.4 & 53.1 \\
\quad \quad w/ 50\% noise & Llama-3 8B (orig) & 42.7 & 47.3 & 50.2 \\
\quad \quad w/ 70\% noise & Llama-3 8B (orig) & 41.9 & 46.3 & 49.6 \\
\bottomrule
\end{tabular}
\end{minipage}
\begin{minipage}{0.35\textwidth}
\begin{tabular}{l c}
\toprule
   \textbf{Ours vs} & \textbf{Score} \\
   \midrule
   Llama-3 8B w/ prompt choice 1  & 0.99 \\
   Llama-3 8B w/ prompt choice 2  & 0.98 \\
   Mistral 8B ~\cite{mistral8b} & 0.98 \\
   \bottomrule
\end{tabular}
\end{minipage}

\label{tab:llm_noise}
\end{table}
}

\subsection{Additional quantitative results}
\label{supp:quant}

In Sec.~\ref{sec:pretraining-results} and Tab.~\ref{tab:merged-results} (top left), we show results for generating skill-attributes for IoU$@0.7$. We extend the table to show results on IoU$@k$ for $k \in \{0.7, 0.8, 1.0\}$ in Tab.~\ref{tab:sa-complete}.

Next, we show zero-shot transfer performance on the QEVD~\cite{qevd-panchal} dataset (summarized in main paper, and detailed here due to space limitations). Note that all videos in QEVD are for fitness exercises, and they are not as distinct as a different sport. Moreover, every video contains multiple exercises, with the transition labeled as \emph{``Moving to \textsc{(exercise name)}...''}. We use these labels to split the videos per-exercise. We discard instances that are before the start of any labeled exercise. Next, we create sport labels 
based on similarity in execution and effects. We create this division for the purpose of zero-shot transfer experiments. This split is created using consensus from ChatGPT-4o~\cite{gpt4o} and Llama-3.1~\cite{llama3modelcard}, and finally manually verified. The splits and the reasonings are given in Tab.~\ref{tab:exercise-groups}. A total of $23$ unique exercises are divided into $5$ groups.

Fig.~\ref{fig:zs-qevd} shows the results. First, in skill-attribute generation (top row), we see that our method outperforms both Stream-VLM~\cite{qevd-panchal} and InternVideo2~\cite{internvideo2} baselines. Moreover, as seen in Ego-Exo4D~\cite{egoexo4d}, the performance decrease in the zero-shot setting is milder than the drop observed in the baselines. Finally, we see a similar trend in actionable feedback generation (bottom row). Overall, zero-shot results in both Ego-Exo4D~\cite{egoexo4d} and QEVD~\cite{qevd-panchal} show that our idea of learning to generalize using skill-attributes is effective.

\KACr{Finally, to show the robustness and sensitivity, we perform the following experiments on Ego-Exo4D~\cite{egoexo4d} actionable feedback generation:

\emph{Robustness to the choice of the language model}: We train and evaluate skill-attributes using different language models. We first train the model using Mistral’s 8B language model (mistralai/Ministral-8B-Instruct-2410)~\cite{mistral8b} and compare it against the skill-attributes test set obtained using Llama-3.1-8B-Instruct, and vice-versa. Tab.~\ref{tab:llm_noise} (left) shows the results. We see that the model is robust to the choice of the language model, and using any strong language model helps achieve a good performance.

\emph{Actionable feedback generation w/ noisy skill-attributes}: We inject noise in the actionable feedback generation evaluation. We replace X\% of inferred skill-attributes with a random skill-attribute and observe the performance at various levels of noise X. See results in the table below. We observe that adding noise degrades the performance, with the performance matching that of end-to-end direct training at $X=20\%$. We can conclude that the performance is positively correlated to the quality of the generated skill-attribute. Improving that will also improve the actionable feedback performance. These ablation studies further showcase the effectiveness of using skill-attributes for actionable feedback.

\emph{Correlation between skill-attributes generated using different LLMs}: We try two prompt variants, and a different language model (Mistral 8B, mistralai/Ministral-8B-Instruct-2410)~\cite{mistral8b} to compare the similarity between generated skill-attributes---checking if our idea is independent of the chosen language model. We use Hungarian matching to find the most-similar match between the new skill-attribute set and our original skill-attribute set. Next, we find the average BERT score~\cite{bert} between the sets. Tab~\ref{tab:llm_noise} (right) shows the results.
We see a very high similarity between skill-attributes generated from different prompts, and a different language model. This result implies that our idea is independent of the choice of a reasonable language model.

}

\subsection{Limitations}
\label{supp:limitations}

We observe that the proposed model struggles with feedback that is about aspects not directly visible in the video. Phrases like \emph{lacking intent} is not groundable definitively and hence, is not captured. Nevertheless, this inability to capture abstract notions is also observable in all the baselines, and in general, vision encoders. Secondly, as we discuss in Sec.~\ref{sec:af-results}, commentary is subjective and there can be various correct ways of providing feedback that improves a learner's performance.

\KACr{Moreover, in this work, we do not factor aspects like terrain and opponent behavior. This assumption is consistent with the datasets Ego-Exo4D~\cite{egoexo4d} and QEVD~\cite{qevd-panchal} that are both single-person, and do not consider external factors. Furthermore, prior work also considers single-person skill assessment/feedback~\cite{expertaf,baller-gedas,egoexo4d}. There are research works in multi-agent cooperation, but they are restricted to simulation and simpler objectives than performance feedback.}

\subsection{Societal impact}
\label{supp:societal}

Our \modelname~can be used for learning skills, especially long-tailed low-resource sports like \emph{kho-kho}, \emph{shinty}. On the positive side, our model democratizes access to skill coaching, and it promotes inclusivity in underrepresented sports. More learning will promote more people playing the sport, and eventually, more data for training and expansion of knowledge. However, the model is trained with Ego-Exo4D~\cite{egoexo4d} and QEVD~\cite{qevd-panchal} that might have regional bias. We believe as more data is available, the biases will go down, and we will move closer towards full physical skill understanding.

\clearpage

%% file: sec/Y_checklist.tex
\newpage
\section*{NeurIPS Paper Checklist}

The checklist is designed to encourage best practices for responsible machine learning research, addressing issues of reproducibility, transparency, research ethics, and societal impact. Do not remove the checklist: {\bf The papers not including the checklist will be desk rejected.} The checklist should follow the references and follow the (optional) supplemental material.  The checklist does NOT count towards the page
limit. 

Please read the checklist guidelines carefully for information on how to answer these questions. For each question in the checklist:
\begin{itemize}
    \item You should answer \answerYes{}, \answerNo{}, or \answerNA{}.
    \item \answerNA{} means either that the question is Not Applicable for that particular paper or the relevant information is Not Available.
    \item Please provide a short (1–2 sentence) justification right after your answer (even for NA). 
\end{itemize}

{\bf The checklist answers are an integral part of your paper submission.} They are visible to the reviewers, area chairs, senior area chairs, and ethics reviewers. You will be asked to also include it (after eventual revisions) with the final version of your paper, and its final version will be published with the paper.

The reviewers of your paper will be asked to use the checklist as one of the factors in their evaluation. While "\answerYes{}" is generally preferable to "\answerNo{}", it is perfectly acceptable to answer "\answerNo{}" provided a proper justification is given (e.g., "error bars are not reported because it would be too computationally expensive" or "we were unable to find the license for the dataset we used"). In general, answering "\answerNo{}" or "\answerNA{}" is not grounds for rejection. While the questions are phrased in a binary way, we acknowledge that the true answer is often more nuanced, so please just use your best judgment and write a justification to elaborate. All supporting evidence can appear either in the main paper or the supplemental material, provided in appendix. If you answer \answerYes{} to a question, in the justification please point to the section(s) where related material for the question can be found.

IMPORTANT, please:
\begin{itemize}
    \item {\bf Delete this instruction block, but keep the section heading ``NeurIPS Paper Checklist"},
    \item  {\bf Keep the checklist subsection headings, questions/answers and guidelines below.}
    \item {\bf Do not modify the questions and only use the provided macros for your answers}.
\end{itemize}

\begin{enumerate}

\item {\bf Claims}
    \item[] Question: Do the main claims made in the abstract and introduction accurately reflect the paper's contributions and scope?
    \item[] Answer: \answerYes{} %
    \item[] Justification: \answerYes{We show results on transferring performance across various sports on multiple datasets.}
    \item[] Guidelines:
    \begin{itemize}
        \item The answer NA means that the abstract and introduction do not include the claims made in the paper.
        \item The abstract and/or introduction should clearly state the claims made, including the contributions made in the paper and important assumptions and limitations. A No or NA answer to this question will not be perceived well by the reviewers. 
        \item The claims made should match theoretical and experimental results, and reflect how much the results can be expected to generalize to other settings. 
        \item It is fine to include aspirational goals as motivation as long as it is clear that these goals are not attained by the paper. 
    \end{itemize}

\item {\bf Limitations}
    \item[] Question: Does the paper discuss the limitations of the work performed by the authors?
    \item[] Answer: \answerYes{} %
    \item[] Justification: \answerYes{Sec.~\ref{sec:youtube-results} discusses the limitations of our work.}
    \item[] Guidelines:
    \begin{itemize}
        \item The answer NA means that the paper has no limitation while the answer No means that the paper has limitations, but those are not discussed in the paper. 
        \item The authors are encouraged to create a separate "Limitations" section in their paper.
        \item The paper should point out any strong assumptions and how robust the results are to violations of these assumptions (e.g., independence assumptions, noiseless settings, model well-specification, asymptotic approximations only holding locally). The authors should reflect on how these assumptions might be violated in practice and what the implications would be.
        \item The authors should reflect on the scope of the claims made, e.g., if the approach was only tested on a few datasets or with a few runs. In general, empirical results often depend on implicit assumptions, which should be articulated.
        \item The authors should reflect on the factors that influence the performance of the approach. For example, a facial recognition algorithm may perform poorly when image resolution is low or images are taken in low lighting. Or a speech-to-text system might not be used reliably to provide closed captions for online lectures because it fails to handle technical jargon.
        \item The authors should discuss the computational efficiency of the proposed algorithms and how they scale with dataset size.
        \item If applicable, the authors should discuss possible limitations of their approach to address problems of privacy and fairness.
        \item While the authors might fear that complete honesty about limitations might be used by reviewers as grounds for rejection, a worse outcome might be that reviewers discover limitations that aren't acknowledged in the paper. The authors should use their best judgment and recognize that individual actions in favor of transparency play an important role in developing norms that preserve the integrity of the community. Reviewers will be specifically instructed to not penalize honesty concerning limitations.
    \end{itemize}

\item {\bf Theory assumptions and proofs}
    \item[] Question: For each theoretical result, does the paper provide the full set of assumptions and a complete (and correct) proof?
    \item[] Answer: \answerNA{} %
    \item[] Justification: \answerNA{We do not propose any theoretical result.}
    \item[] Guidelines:
    \begin{itemize}
        \item The answer NA means that the paper does not include theoretical results. 
        \item All the theorems, formulas, and proofs in the paper should be numbered and cross-referenced.
        \item All assumptions should be clearly stated or referenced in the statement of any theorems.
        \item The proofs can either appear in the main paper or the supplemental material, but if they appear in the supplemental material, the authors are encouraged to provide a short proof sketch to provide intuition. 
        \item Inversely, any informal proof provided in the core of the paper should be complemented by formal proofs provided in appendix or supplemental material.
        \item Theorems and Lemmas that the proof relies upon should be properly referenced. 
    \end{itemize}

    \item {\bf Experimental result reproducibility}
    \item[] Question: Does the paper fully disclose all the information needed to reproduce the main experimental results of the paper to the extent that it affects the main claims and/or conclusions of the paper (regardless of whether the code and data are provided or not)?
    \item[] Answer: \answerYes{} %
    \item[] Justification: \answerYes{Sec.~\ref{sec:training-and-inference} describes all the information required to reproduce the code. Furthermore, we will release the code and the data upon paper acceptance.}
    \item[] Guidelines:
    \begin{itemize}
        \item The answer NA means that the paper does not include experiments.
        \item If the paper includes experiments, a No answer to this question will not be perceived well by the reviewers: Making the paper reproducible is important, regardless of whether the code and data are provided or not.
        \item If the contribution is a dataset and/or model, the authors should describe the steps taken to make their results reproducible or verifiable. 
        \item Depending on the contribution, reproducibility can be accomplished in various ways. For example, if the contribution is a novel architecture, describing the architecture fully might suffice, or if the contribution is a specific model and empirical evaluation, it may be necessary to either make it possible for others to replicate the model with the same dataset, or provide access to the model. In general. releasing code and data is often one good way to accomplish this, but reproducibility can also be provided via detailed instructions for how to replicate the results, access to a hosted model (e.g., in the case of a large language model), releasing of a model checkpoint, or other means that are appropriate to the research performed.
        \item While NeurIPS does not require releasing code, the conference does require all submissions to provide some reasonable avenue for reproducibility, which may depend on the nature of the contribution. For example
        \begin{enumerate}
            \item If the contribution is primarily a new algorithm, the paper should make it clear how to reproduce that algorithm.
            \item If the contribution is primarily a new model architecture, the paper should describe the architecture clearly and fully.
            \item If the contribution is a new model (e.g., a large language model), then there should either be a way to access this model for reproducing the results or a way to reproduce the model (e.g., with an open-source dataset or instructions for how to construct the dataset).
            \item We recognize that reproducibility may be tricky in some cases, in which case authors are welcome to describe the particular way they provide for reproducibility. In the case of closed-source models, it may be that access to the model is limited in some way (e.g., to registered users), but it should be possible for other researchers to have some path to reproducing or verifying the results.
        \end{enumerate}
    \end{itemize}

\item {\bf Open access to data and code}
    \item[] Question: Does the paper provide open access to the data and code, with sufficient instructions to faithfully reproduce the main experimental results, as described in supplemental material?
    \item[] Answer: \answerNo{} %
    \item[] Justification: \answerNo{The code and data will be released upon acceptance.}
    \item[] Guidelines:
    \begin{itemize}
        \item The answer NA means that paper does not include experiments requiring code.
        \item Please see the NeurIPS code and data submission guidelines (\url{https://nips.cc/public/guides/CodeSubmissionPolicy}) for more details.
        \item While we encourage the release of code and data, we understand that this might not be possible, so “No” is an acceptable answer. Papers cannot be rejected simply for not including code, unless this is central to the contribution (e.g., for a new open-source benchmark).
        \item The instructions should contain the exact command and environment needed to run to reproduce the results. See the NeurIPS code and data submission guidelines (\url{https://nips.cc/public/guides/CodeSubmissionPolicy}) for more details.
        \item The authors should provide instructions on data access and preparation, including how to access the raw data, preprocessed data, intermediate data, and generated data, etc.
        \item The authors should provide scripts to reproduce all experimental results for the new proposed method and baselines. If only a subset of experiments are reproducible, they should state which ones are omitted from the script and why.
        \item At submission time, to preserve anonymity, the authors should release anonymized versions (if applicable).
        \item Providing as much information as possible in supplemental material (appended to the paper) is recommended, but including URLs to data and code is permitted.
    \end{itemize}

\item {\bf Experimental setting/details}
    \item[] Question: Does the paper specify all the training and test details (e.g., data splits, hyperparameters, how they were chosen, type of optimizer, etc.) necessary to understand the results?
    \item[] Answer: \answerYes{}{} %
    \item[] Justification: \answerYes{The method section (Sec. \ref{sec:method}) discusses the training and test details, along with all the design choices.}
    \item[] Guidelines:
    \begin{itemize}
        \item The answer NA means that the paper does not include experiments.
        \item The experimental setting should be presented in the core of the paper to a level of detail that is necessary to appreciate the results and make sense of them.
        \item The full details can be provided either with the code, in appendix, or as supplemental material.
    \end{itemize}

\item {\bf Experiment statistical significance}
    \item[] Question: Does the paper report error bars suitably and correctly defined or other appropriate information about the statistical significance of the experiments?
    \item[] Answer: \answerYes{} %
    \item[] Justification: \answerYes{All result sections have the maximum standard error reported.}
    \item[] Guidelines:
    \begin{itemize}
        \item The answer NA means that the paper does not include experiments.
        \item The authors should answer "Yes" if the results are accompanied by error bars, confidence intervals, or statistical significance tests, at least for the experiments that support the main claims of the paper.
        \item The factors of variability that the error bars are capturing should be clearly stated (for example, train/test split, initialization, random drawing of some parameter, or overall run with given experimental conditions).
        \item The method for calculating the error bars should be explained (closed form formula, call to a library function, bootstrap, etc.)
        \item The assumptions made should be given (e.g., Normally distributed errors).
        \item It should be clear whether the error bar is the standard deviation or the standard error of the mean.
        \item It is OK to report 1-sigma error bars, but one should state it. The authors should preferably report a 2-sigma error bar than state that they have a 96\% CI, if the hypothesis of Normality of errors is not verified.
        \item For asymmetric distributions, the authors should be careful not to show in tables or figures symmetric error bars that would yield results that are out of range (e.g. negative error rates).
        \item If error bars are reported in tables or plots, The authors should explain in the text how they were calculated and reference the corresponding figures or tables in the text.
    \end{itemize}

\item {\bf Experiments compute resources}
    \item[] Question: For each experiment, does the paper provide sufficient information on the computer resources (type of compute workers, memory, time of execution) needed to reproduce the experiments?
    \item[] Answer: \answerYes{} %
    \item[] Justification: \answerYes{The details of compute resource and training duration is given in Sec.~\ref{sec:training-and-inference}. We did not have significant computational overhead due to failed experiments.}
    \item[] Guidelines:
    \begin{itemize}
        \item The answer NA means that the paper does not include experiments.
        \item The paper should indicate the type of compute workers CPU or GPU, internal cluster, or cloud provider, including relevant memory and storage.
        \item The paper should provide the amount of compute required for each of the individual experimental runs as well as estimate the total compute. 
        \item The paper should disclose whether the full research project required more compute than the experiments reported in the paper (e.g., preliminary or failed experiments that didn't make it into the paper). 
    \end{itemize}
    
\item {\bf Code of ethics}
    \item[] Question: Does the research conducted in the paper conform, in every respect, with the NeurIPS Code of Ethics \url{https://neurips.cc/public/EthicsGuidelines}?
    \item[] Answer: \answerYes{} %
    \item[] Justification: \answerYes{The authors confirm that this research follows the NeurIPS code of ethics.}
    \item[] Guidelines:
    \begin{itemize}
        \item The answer NA means that the authors have not reviewed the NeurIPS Code of Ethics.
        \item If the authors answer No, they should explain the special circumstances that require a deviation from the Code of Ethics.
        \item The authors should make sure to preserve anonymity (e.g., if there is a special consideration due to laws or regulations in their jurisdiction).
    \end{itemize}

\item {\bf Broader impacts}
    \item[] Question: Does the paper discuss both potential positive societal impacts and negative societal impacts of the work performed?
    \item[] Answer: \answerYes{} %
    \item[] Justification: \answerYes{Supplementary discusses societal impact.}
    \item[] Guidelines:
    \begin{itemize}
        \item The answer NA means that there is no societal impact of the work performed.
        \item If the authors answer NA or No, they should explain why their work has no societal impact or why the paper does not address societal impact.
        \item Examples of negative societal impacts include potential malicious or unintended uses (e.g., disinformation, generating fake profiles, surveillance), fairness considerations (e.g., deployment of technologies that could make decisions that unfairly impact specific groups), privacy considerations, and security considerations.
        \item The conference expects that many papers will be foundational research and not tied to particular applications, let alone deployments. However, if there is a direct path to any negative applications, the authors should point it out. For example, it is legitimate to point out that an improvement in the quality of generative models could be used to generate deepfakes for disinformation. On the other hand, it is not needed to point out that a generic algorithm for optimizing neural networks could enable people to train models that generate Deepfakes faster.
        \item The authors should consider possible harms that could arise when the technology is being used as intended and functioning correctly, harms that could arise when the technology is being used as intended but gives incorrect results, and harms following from (intentional or unintentional) misuse of the technology.
        \item If there are negative societal impacts, the authors could also discuss possible mitigation strategies (e.g., gated release of models, providing defenses in addition to attacks, mechanisms for monitoring misuse, mechanisms to monitor how a system learns from feedback over time, improving the efficiency and accessibility of ML).
    \end{itemize}
    
\item {\bf Safeguards}
    \item[] Question: Does the paper describe safeguards that have been put in place for responsible release of data or models that have a high risk for misuse (e.g., pretrained language models, image generators, or scraped datasets)?
    \item[] Answer: \answerYes{} %
    \item[] Justification: \answerYes{We only use Exo-Exo4D~\cite{egoexo4d} 
 and QEVD~\cite{qevd-panchal} dataset for the research. Moreover, the generations of the language model is finetuned to only generate expert commentary. We believe our trained model cannot be misused.}
    \item[] Guidelines:
    \begin{itemize}
        \item The answer NA means that the paper poses no such risks.
        \item Released models that have a high risk for misuse or dual-use should be released with necessary safeguards to allow for controlled use of the model, for example by requiring that users adhere to usage guidelines or restrictions to access the model or implementing safety filters. 
        \item Datasets that have been scraped from the Internet could pose safety risks. The authors should describe how they avoided releasing unsafe images.
        \item We recognize that providing effective safeguards is challenging, and many papers do not require this, but we encourage authors to take this into account and make a best faith effort.
    \end{itemize}

\item {\bf Licenses for existing assets}
    \item[] Question: Are the creators or original owners of assets (e.g., code, data, models), used in the paper, properly credited and are the license and terms of use explicitly mentioned and properly respected?
    \item[] Answer: \answerYes{} %
    \item[] Justification: \answerYes{We have cited and ensured all the codes and models have licenses that we can use for this research.}
    \item[] Guidelines:
    \begin{itemize}
        \item The answer NA means that the paper does not use existing assets.
        \item The authors should cite the original paper that produced the code package or dataset.
        \item The authors should state which version of the asset is used and, if possible, include a URL.
        \item The name of the license (e.g., CC-BY 4.0) should be included for each asset.
        \item For scraped data from a particular source (e.g., website), the copyright and terms of service of that source should be provided.
        \item If assets are released, the license, copyright information, and terms of use in the package should be provided. For popular datasets, \url{paperswithcode.com/datasets} has curated licenses for some datasets. Their licensing guide can help determine the license of a dataset.
        \item For existing datasets that are re-packaged, both the original license and the license of the derived asset (if it has changed) should be provided.
        \item If this information is not available online, the authors are encouraged to reach out to the asset's creators.
    \end{itemize}

\item {\bf New assets}
    \item[] Question: Are new assets introduced in the paper well documented and is the documentation provided alongside the assets?
    \item[] Answer: \answerYes{} %
    \item[] Justification: \answerYes{All the details of the weakly-supervised dataset is provided in Sec. \ref{sec:method}. We will provide the documentation of the code along with its release to the community.}
    \item[] Guidelines:
    \begin{itemize}
        \item The answer NA means that the paper does not release new assets.
        \item Researchers should communicate the details of the dataset/code/model as part of their submissions via structured templates. This includes details about training, license, limitations, etc. 
        \item The paper should discuss whether and how consent was obtained from people whose asset is used.
        \item At submission time, remember to anonymize your assets (if applicable). You can either create an anonymized URL or include an anonymized zip file.
    \end{itemize}

\item {\bf Crowdsourcing and research with human subjects}
    \item[] Question: For crowdsourcing experiments and research with human subjects, does the paper include the full text of instructions given to participants and screenshots, if applicable, as well as details about compensation (if any)? 
    \item[] Answer: \answerNA{} %
    \item[] Justification: \answerNA{}
    \item[] Guidelines:
    \begin{itemize}
        \item The answer NA means that the paper does not involve crowdsourcing nor research with human subjects.
        \item Including this information in the supplemental material is fine, but if the main contribution of the paper involves human subjects, then as much detail as possible should be included in the main paper. 
        \item According to the NeurIPS Code of Ethics, workers involved in data collection, curation, or other labor should be paid at least the minimum wage in the country of the data collector. 
    \end{itemize}

\item {\bf Institutional review board (IRB) approvals or equivalent for research with human subjects}
    \item[] Question: Does the paper describe potential risks incurred by study participants, whether such risks were disclosed to the subjects, and whether Institutional Review Board (IRB) approvals (or an equivalent approval/review based on the requirements of your country or institution) were obtained?
    \item[] Answer: \answerNA{} %
    \item[] Justification: \answerNA{}
    \item[] Guidelines:
    \begin{itemize}
        \item The answer NA means that the paper does not involve crowdsourcing nor research with human subjects.
        \item Depending on the country in which research is conducted, IRB approval (or equivalent) may be required for any human subjects research. If you obtained IRB approval, you should clearly state this in the paper. 
        \item We recognize that the procedures for this may vary significantly between institutions and locations, and we expect authors to adhere to the NeurIPS Code of Ethics and the guidelines for their institution. 
        \item For initial submissions, do not include any information that would break anonymity (if applicable), such as the institution conducting the review.
    \end{itemize}

\item {\bf Declaration of LLM usage}
    \item[] Question: Does the paper describe the usage of LLMs if it is an important, original, or non-standard component of the core methods in this research? Note that if the LLM is used only for writing, editing, or formatting purposes and does not impact the core methodology, scientific rigorousness, or originality of the research, declaration is not required.
    \item[] Answer: \answerYes{} %
    \item[] Justification: \answerYes{Our model design use LLM as the core component. This is a standard in recent video understanding methods~\cite{llava,videochat}.}
    \item[] Guidelines:
    \begin{itemize}
        \item The answer NA means that the core method development in this research does not involve LLMs as any important, original, or non-standard components.
        \item Please refer to our LLM policy (\url{https://neurips.cc/Conferences/2025/LLM}) for what should or should not be described.
    \end{itemize}

\end{enumerate}